\documentclass[letterpaper]{article} % DO NOT CHANGE THIS
\usepackage{aaai25}  % DO NOT CHANGE THIS
\usepackage{times}  % DO NOT CHANGE THIS
\usepackage{helvet}  % DO NOT CHANGE THIS
\usepackage{courier}  % DO NOT CHANGE THIS
\usepackage[hyphens]{url}  % DO NOT CHANGE TH
\usepackage{graphicx} % DO NOT CHANGE THIS
\usepackage{natbib}  % DO NOT CHANGE THIS AND DO NOT ADD ANY OPTIONS TO IT
\usepackage{caption} % DO NOT CHANGE THIS AND DO NOT ADD ANY OPTIONS TO IT
\usepackage{algorithm}
\usepackage{algorithmic}

\usepackage{newfloat}
\usepackage{listings}
\DeclareCaptionStyle{ruled}{labelfont=normalfont,labelsep=colon,strut=off} % DO NOT CHANGE THIS
\floatstyle{ruled}
\newfloat{listing}{tb}{lst}{}
\floatname{listing}{Listing}
\newcounter{eqfncorr}
\def\equalcorr{
  \ifnum\value{eqfncorr}=0
    \footnote{Corresponding Authors}
    \setcounter{eqfncorr}{\value{footnote}}
  \else
    \footnotemark[\value{eqfncorr}]
  \fi
}

\title{DualCP: Rehearsal-Free Domain-Incremental Learning \\
via Dual-Level Concept Prototype}

\author{Qiang Wang\textsuperscript{\rm 1},
Yuhang He\textsuperscript{\rm 1}\equalcorr,
Songlin Dong\textsuperscript{\rm 1}, 
Xiang Song\textsuperscript{\rm 1}, \\
Jizhou Han\textsuperscript{\rm 1}, 
Haoyu Luo\textsuperscript{\rm 1}, 
Yihong Gong\textsuperscript{\rm 1,2}\equalcorr}
\affiliations{
    \textsuperscript{\rm 1}Xi’an Jiaotong University\\
    \textsuperscript{\rm 2}Shenzhen University of Advanced Technology\\
    qwang@stu.xjtu.edu.cn, 
    heyuhang@xjtu.edu.cn, \\
    \{dsl972731417, songxiang, jizhou\_han, luohaoyu\}@stu.xjtu.edu.cn, 
    ygong@mail.xjtu.edu.cn
}

\usepackage{amsmath}
\usepackage{amssymb}
\usepackage[table]{xcolor}
\usepackage{booktabs}
\usepackage{multirow}
\usepackage{arydshln}
\usepackage{pifont}
\usepackage{soul}
\usepackage[capitalize]{cleveref}
\usepackage{diagbox}

\begin{document}

\maketitle

\begin{abstract}
Domain-Incremental Learning (DIL) enables vision models to adapt to changing conditions in real-world environments while maintaining the knowledge acquired from previous domains. Given privacy concerns and training time, Rehearsal-Free DIL (RFDIL) is more practical. Inspired by the incremental cognitive process of the human brain, we design Dual-level Concept Prototypes (DualCP) for each class to address the conflict between learning new knowledge and retaining old knowledge in RFDIL. To construct DualCP, we propose a Concept Prototype Generator (CPG) that generates both coarse-grained and fine-grained prototypes for each class. Additionally, we introduce a Coarse-to-Fine calibrator (C2F) to align image features with DualCP. Finally, we propose a Dual Dot-Regression (DDR) loss function to optimize our C2F module. Extensive experiments on the DomainNet, CDDB, and CORe50 datasets demonstrate the effectiveness of our method.
\end{abstract}

\section{Introduction}

Domain-Incremental Learning (DIL) aims to train a unified model incrementally across continuously encountered domains. It has drawn remarkable attention in recent years~\cite{zhang2022claire,verwimp2023clad} and has a wide range of applications. For example, in a visual recognition model on an autonomous vehicle, the model is expected to incrementally learn and adapt to new and dynamic environments~\cite{wang2024enhancing} such as forests, deserts, cities, \emph{etc}. As the number of domains increases, the model may forget previously acquired knowledge. Many studies~\cite{isele2018selective,zhao2021memory} address the retention of knowledge by preserving and retraining old samples, known as \emph{rehearsal}. Rehearsal-based methods, however, require longer training time, and storing old domain images may raise privacy concerns~\cite{wan2024prompt}. Therefore, Rehearsal-Free DIL (RFDIL) is more practical for real-world scenarios compared to rehearsal-based approaches.

\begin{figure}[t]
  \centering
  \includegraphics[width=0.45\textwidth]{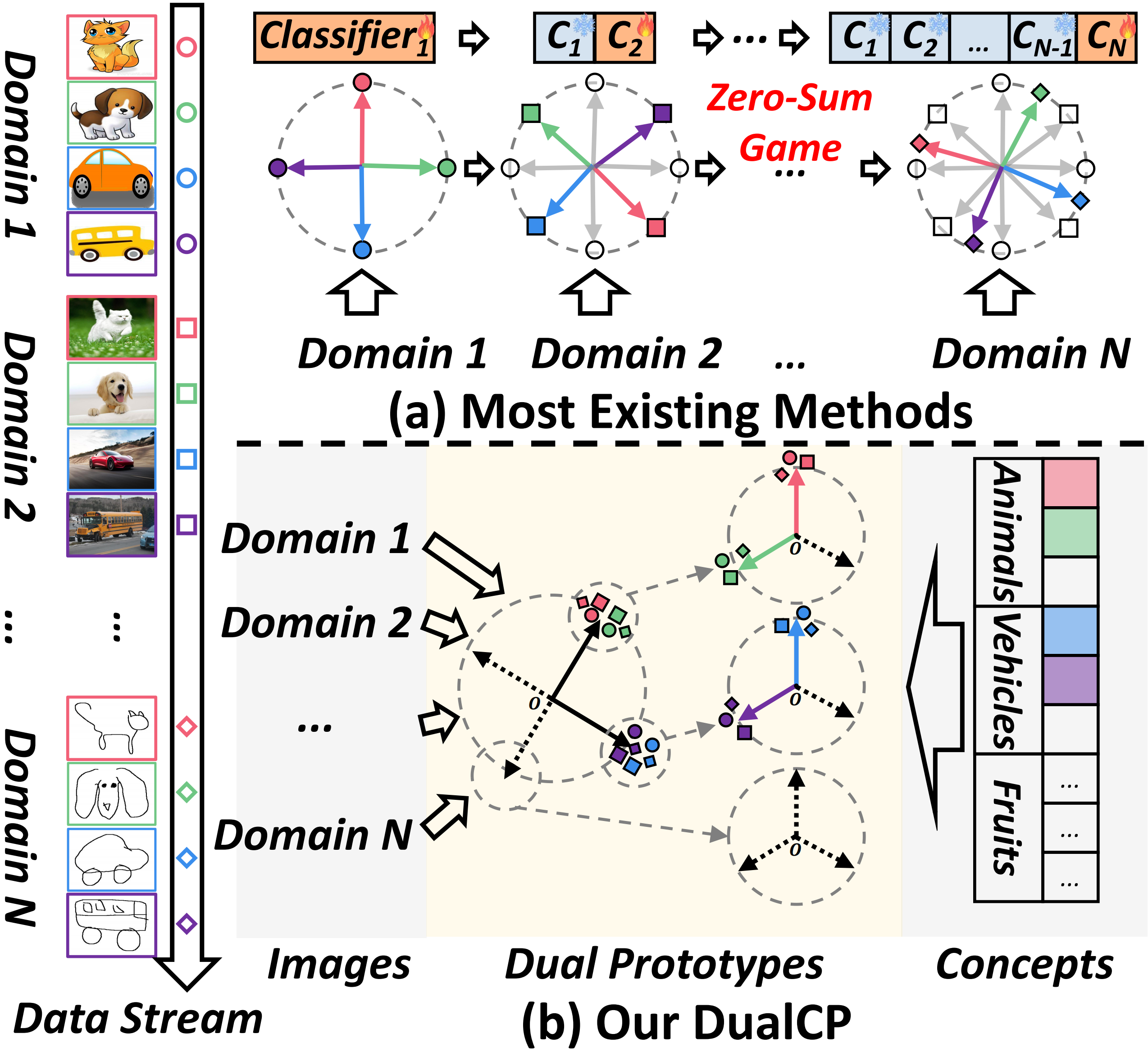}
  \caption{(a) Illustration of most existing methods, where new domain features and old domain features compete within the feature space. (b) Illustration of our method based on dual-level concept prototypes. Different colors (red, green, blue, and purple) represent different classes, while different shapes ($\circ$, $\square$, and $\diamond$) represent different domains. Best viewed in color.}
  \label{fig:introfig}
\end{figure}

The primary challenge of RFDIL is the conflict between learning new domains and preventing forgetting of old ones. On the one hand, learning new domain knowledge may overwrite parameters related to old domains, leading to \emph{catastrophic forgetting}~\cite{mccloskey1989catastrophic}. On the other hand, adding artificial constraints to prevent forgetting old domains will hinder learning new ones. To mitigate this dilemma, the method~\cite{zhu2021prototype} proposes generating samples for the old domains as a supplement to retain knowledge; the methods~\cite{rebuffi2017icarl,hou2019learning} suggest employing knowledge distillation strategies to slow down the forgetting of old knowledge; the methods~\cite{douillard2022dytox,wang2022learning,smith2023coda,wang2022s} propose to train a set of learnable prompt tokens for knowledge retention. However, most methods aim to balance new-domain learning and old-domain forgetting, resulting in a zero-sum game. As illustrated in \cref{fig:introfig}~(a), the increase in the number of domains causes more features to crowd within the same feature space, escalating the competition between new and old domains. This raises a question: \emph{are different domains necessarily in opposition?}

To address this question, we draw inspiration from the incremental cognitive process of humans. 
The cognitive science research~\cite{bar2003cortical,fenske2006top} indicate that the early visual areas of the brain extract both low and high spatial frequency signals (LF \& HF) from the visual stimuli after encountering a new object. 
The LF signals are projected to the prefrontal cortex, forming coarse-grained features within approximately 140 ms~\cite{barcelo2000prefrontal}. For example, when seeing a photograph or a cartoon image of a cat, the brain initially forms a general impression of an animal, without determining the cat’s color or whether the cat is real. This process occurs so quickly that we are often unaware of it in everyday life. 
The HF signals are processed by the ventral stream to generate fine-grained features, such as the texture of the image (identifying the domain it belongs to) or the precise category of the object (\emph{e.g.}, cat or dog within the animal category). 
From this research, we derive two key insights:
\textbf{(a) The learning of new knowledge is hierarchical.} The brain first forms a general concept of a new object before focusing on its details. In contrast, existing deep classification models treat all given categories equally, without considering their inherent semantic relationships. 
\textbf{(b) New and old domains are not completely distinct.} Although domain shifts cause visual differences among different domains, the coarse-grained representations of the same concept are similar from a human perspective. \emph{This answers the above question}.

Based on the above insights, we design a framework based on two principles.
\textbf{(a) We propose treating images of the same class across multiple domains as a single concept.} This concept remains consistent regardless of the domain. For instance, the concept of a ``cat'' does not change whether it is represented by photos, cartoons, or simple sketches of cats. During the initial domain training in RFDIL, we employ neural networks to mine the commonalities of the same concept across different domains. Specifically, we use the semantic features of the concept to construct a concept prototype. To maximize the separability between different prototypes, we represent the concept prototype using a semantically guided simplex Equiangular Tight Frame (ETF) through the neural collapse theory. The ETF is a mathematical structure that maximizes the distances between feature pairs, with further details provided in the preliminary section. 
\textbf{(b) We introduce Dual-level Concept Prototypes (DualCP) to model the human cognitive process of recognizing new objects.} we first group all classes based on their superordinate, as shown in \cref{fig:introfig}~(b). For example, cats and dogs belong to the animal group, while cars and buses belong to the vehicle group. First-level concept prototypes are designed for each superordinate concept to simulate the coarse-grained features formed by LF signals in the prefrontal cortex. Within each group, we design second-level concept prototypes to simulate the process by which HF signals in the ventral stream recognize specific classes. 
Finally, we introduce a coarse-to-fine calibrator (C2F) to align image features with their corresponding concept prototypes. A Dual-level Dot Regression (DDR) loss function is employed to optimize the training of the C2F. We conduct extensive experiments on three benchmark datasets, including DomainNet, CORe50, and CDDB, comparing our DualCP with state-of-the-art methods.

%%% 6
Our contributions can be summarized as follows:
\begin{itemize}
\item Inspired by the incremental cognition process of humans, we propose treating cross-domain images of the same class as a single concept. We construct concept prototypes for all domains to prevent a zero-sum game between new and old domains.
\item We introduce DualCP, an RFDIL method based on dual-level concept prototypes, which further enhances the separability between classes by performing fine-grained classification on similar categories.
\item We propose the C2F module and the DDR loss function to align objects with their corresponding coarse-grained and fine-grained concept prototypes.
\item Extensive experiments demonstrate that the proposed DualCP method outperforms other RFDIL methods on three benchmark datasets, with a margin of up to 7.6\% on the CDDB dataset.
\end{itemize}

\section{Related Work}

\subsection{Domain-Incremental Learning}
Multiple studies focus on addressing \emph{catastrophic forgetting}~\cite{mccloskey1989catastrophic} in Domain-Incremental Learning (DIL). For instance, \cite{kirkpatrick2017overcoming,akyurek2021subspace,shi2023multi} restrict the model's plasticity to balance the learning of new domains with the forgetting of old ones, resulting in a zero-sum game between new and old domains. \textbf{Rehearsal-based} methods~\cite{rebuffi2017icarl,chaudhry2019tiny} mitigate the forgetting of old domains by replaying a subset of representative samples from the old domains, which raises privacy concerns. \textbf{Rehearsal-free} methods~\cite{van2021class,wang2023non,wan2024grid} use synthetic data replay from old domains to reduce forgetting. \cite{wang2022learning,liu2024compositional,gao2024beyond,wang2025non} dynamically expand the model to incrementally store knowledge from new domains, aiming to separate the learning of new and old domain knowledge to avoid conflicts.

\subsection{Neural Collapse}
The Neural Collapse (NC) theory~\cite{papyan2020prevalence} posits that features of the same class in the final layer collapse to a single point on a hypersphere in a well-trained classification model. Besides, the distance between features of any two different classes is maximized, and these features of all classes collectively can be defined by a simplex Equiangular Tight Frame (ETF). This indicates that a ``training endpoint'' can be easily constructed for the model. We can align image features to the pre-designed endpoint, which is independent of the model's initial parameters. The papers~\cite{mixon2020neural,ji2021unconstrained,zhou2022optimization} have shown that the accuracy of models trained using the simplex ETF is comparable to those trained with conventional cross-entropy loss.

\subsection{Applications of Neural Collapse in Other Tasks}
Recent studies have utilized the above characteristics of NC to address various tasks, such as imbalanced learning~\cite{yang2022inducing}, few-shot class-incremental learning~\cite{yang2023neural}, and federated learning~\cite{huang2023neural}.
Compared with existing research, the proposed DualCP is \textbf{not a simple combination of NC and RFDIL, but is a novel method inspired by the cognitive science research}, \emph{i.e.}, the learning of new knowledge is hierarchical and the new and old domains are not completely distinct. We build concept prototypes based on the two insights to solve the RFDIL problem, while the ETF serves as a tool to realize our idea.

For the classification of a large number of classes, \cite{jiang2023generalized} proposed maximizing the minimum one-vs-rest margins to achieve generalized neural collapse. However, this method has overlooked the inherent similarities among classes. Our proposed DualCP instead groups the numerous classes using text features, then performs coarse-grained classification between groups and fine-grained classification within groups. Our proposed dual-level concept prototypes are not only easy to implement but also accommodate up to $d^{2}$ classes when the feature dimension is $d$, which is sufficient for most real-world scenarios.

\section{Preliminary}

\subsection{Problem Formulation}
Rehearsal-Free Domain-Incremental Learning (RFDIL) aims to train a unified model progressively on the data from $T$ domains $\mathcal{D}=\{\mathcal{D}_{t}\}_{t=1}^{T}$, where $T$ is the total number of domains.
The data in each domain is split into the training set $\mathcal{X}_{t}$ and the test set $\mathcal{Z}_{t}$, denoted as $\mathcal{D}_{t}=(\mathcal{X}_{t},\mathcal{Z}_{t})$.
At the $t$-th stage of training, the model is only allowed to train on $\mathcal{X}_{t}$ and does not have access to $\mathcal{X}_{1 \sim t-1} = \bigcup_{\tau = 1}^{t-1} \mathcal{X}_{\tau}$.
After the $t$-th training, the model is tested on $\mathcal{Z}_{1 \sim t} = \bigcup_{\tau = 1}^{t} \mathcal{Z}_{\tau}$.
The training set $\mathcal{X}_{t}$ is composed of $N_{t}$ tuples, denoted as $\mathcal{X}_{t}=\{(x_{t,i}, y_{t,i})\}_{i=1}^{N_{t}}$, where $x_{t,i}$ signifies the $i$-th image from the $t$-th domain, and  $y_{t,i}$ denotes the label of $x_{t,i}$.
The test set $\mathcal{Z}_{t}$ follows a similar structure.
Moreover, define $\mathcal{C}_{t}$ as the set of labels for the $t$-th domain, \emph{i.e.}, $\forall i \in [1, N_{t}], y_{t,i} \in \mathcal{C}_{t}$.
In other words, $|\mathcal{C}_{t}|$, the number of classes in each domain, remains constant in the same dataset.

\subsection{Definition of Simplex ETF}
For a well-trained classification model with $K$ classes, the within-class means correspond to $K$ prototypes, denoted as $\mathbf{m}_i \in \mathbb{R}^{d}, i=1,2,\cdots,K$, where $K\le d+1$. The collection of these prototypes, \emph{i.e.}, $\mathbf{M} = [\mathbf{m}_1, \cdots, \mathbf{m}_K]$ in $\mathbb{R}^{d \times K}$, is called a simplex ETF, which means:
\begin{equation}
\label{eq:ETF}
  \mathbf{M}=
    \sqrt{\frac{K}{K-1}} 
    \mathbf{U} 
    (\mathbf{I}_{K}-
    \frac{1}{K}
    \mathbf{1}_{K}\mathbf{1}_{K}^{T}), 
\end{equation}
where $\mathbf{U}$ satisfies $\mathbf{U}^{T}\mathbf{U}=\mathbf{I}_{K}$, and $\mathbf{1}_{K}$ is a $K$-dimensional all-ones vector. All prototypes $\mathbf{m}_i$ have the same $l_{2}$ norm, \emph{i.e.}, $|\mathbf{m}_i|=1, i=1,2,\cdots,K$, and the same pair-wise angle, \emph{i.e.}, 
\begin{equation}
\label{eq:ETF_prototype}
  \mathbf{m}_{i}^{T}\mathbf{m}_{j}=
    \frac{K}{K-1} \delta_{i,j}-\frac{1}{K-1},
    \forall i,j \in [1, K],
\end{equation}
where $\delta_{i,j}$ equals to 1 when $i=j$ and 0 otherwise. The pairwise angle $-\frac{1}{K-1} $ is the maximal equiangular separation of $K$ prototypes in the $d$-dimension feature space.

Based on the definition of simplex ETF, the NC phenomenon can be summarized as:

\noindent\textbf{(NC1)} Feature collapse. The last-layer features of the same class will collapse to their within-class mean, \emph{i.e.}, ${\textstyle \sum_{W}^{}} \rightarrow 0$, where ${\textstyle \sum_{W}^{}} = \text{Avg}\{(\mu_{k, i}-\mu_{k})(\mu_{k, i}-\mu_{k})^{T}\}$, $\mu_{k, i}$ is the feature of the $i$-th sample of the $k$-th class and $\mu_{k}$ is the within-class mean of the $k$-th class.

\noindent\textbf{(NC2)} Convergence to simplex ETF. The within-class means of all $K$ classes is centered by the global mean $\mu_{G} = \text{Avg}_{i,k}(\mu_{k,i})$.
These means $\mu_{k}$ will converge to the $K$
prototypes of a simplex ETF, \emph{i.e.}, $\hat{\mu}_{k} = (\mu_{k} - \mu_{G}) / \left \| \mu_{k} - \mu_{G} \right \| , 1 \le k \le K$, where $\hat{\mathbf{M}} = [\hat{\mu}_{1}, \cdots, \hat{\mu}_{K}]$ satisfies \cref{eq:ETF}, and $\hat{\mu}_{k}$ satisfies \cref{eq:ETF_prototype}.

\noindent\textbf{(NC3)} Self-duality. The within-class feature means will be aligned with their corresponding classifier weights $\mathbf{w}_{k}$, \emph{i.e.}, $\hat{\mu}_{k} = \mathbf{w}_{k} / \left \| \mathbf{w}_{k} \right \|$.

\noindent\textbf{(NC4)} Based on {(NC1)-(NC3)}, the model prediction using the classifier can be simplified to select the nearest class center, \emph{i.e.}, $\text{argmax}_{k}\langle \mathbf{\mu}, \mathbf{w}_{k} \rangle = \text{argmin}_{k} \left \| \mu - \mu_{k} \right \|$, where $\langle \cdot \rangle$ is the inner product operator, and $\mu$ is the last-layer feature of a sample for prediction.

\section{Method}

\begin{figure*}[t]
  \centering
  \includegraphics[width=0.9\textwidth]{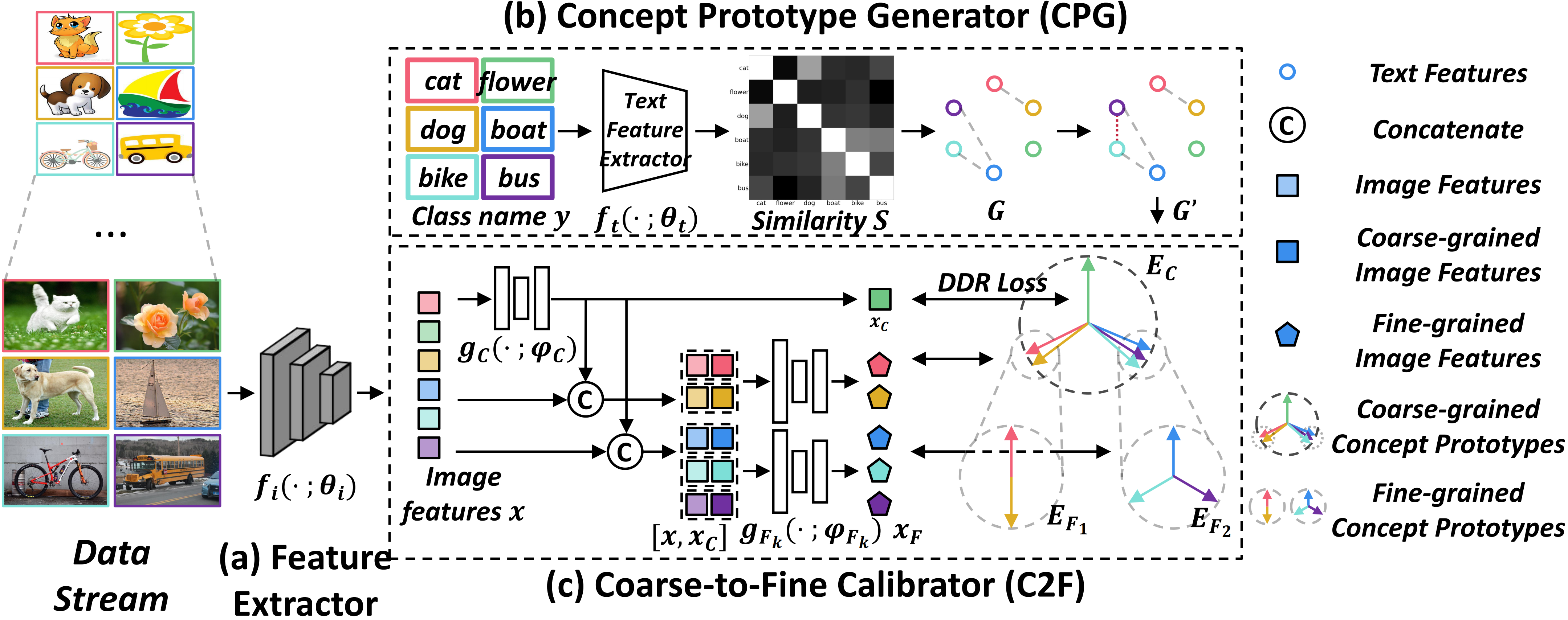}
  \caption{
  \textbf{The framework of the proposed DualCP.} DualCP comprises three main components: (a) a feature extractor to get the image features, (b) the CPG module to construct the dual-level concept prototype based on the text features of the class names, and (c) the C2F module to align the image features with the corresponding prototypes.
  We selected six common classes from the DomainNet dataset, \emph{i.e.}, cat, flower, dog, boat, bike, and bus, to further illustrate our method. Similar classes were grouped, such as cats and dogs. We constructed coarse-grained prototypes between groups and fine-grained prototypes within groups. This coarse-to-fine classification approach helps the model better distinguish similar categories. Best viewed in color.}
  \label{fig:framework}
\end{figure*}

\subsection{Overall Framework}
\cref{fig:framework} illustrates the framework of our proposed DualCP. Our method contains three main components: (a) a pre-trained image feature extractor $f_{i}(\cdot;\theta_{i})$, (b) the Concept Prototype Generator (CPG) with a pre-trained text feature extractor $f_{t}(\cdot;\theta_{t})$, and (c) the Coarse-to-Fine calibrator (C2F), which includes a coarse-grained layer $g_{C}(\cdot;\varphi_{C})$ and multiple fine-grained layers $g_{F_{k}}(\cdot;\varphi_{F_{k}})$.

For simplicity, we abbreviate $x_{t,i}$ as $x$ and $y_{t,i}$ as $y$. Given a set of images $x$ and their class names $y$, we first extract the image features by $\mathbf{x}=f_{i}(x;\theta_{i})$. Then, the class names are fed into the CPG to extract text features and construct dual-level concept prototypes, which include coarse-grained prototypes $\mathbf{E}_{C}$ and fine-grained prototypes $\mathbf{E}_{F_{i}}$. Finally, we train the C2F to align the image features $\mathbf{x}$ with the corresponding concept prototypes by minimizing the proposed Dual Dot-Regression (DDR) loss function.
For further details on the training and inference of our DualCP framework, please refer to the appendix.

\subsection{Concept Prototype Generator}
\paragraph{Preparation for generating prototypes.}
First, we collect the names of all classes, denoted as $\mathcal{C}_{t}=\{y_{1},y_{2},\cdots,y_{K}\}$, \emph{e.g.}, $\mathcal{C}_{t}=\{\text{airplane, bike, cat, ..., zebra}\}$. $K=|\mathcal{C}_{t}|$ is the number of classes. Text features are extracted by:
\begin{align}
  \mathbf{y}_{i} &= \frac{f_{t}(y_{i};\theta_{t})}{\|f_{t}(y_{i};\theta_{t})\|}, \label{eq:text_encoder} \\
  \mathbf{Y} = [&\mathbf{y}_{1}, \mathbf{y}_{2}, \cdots, \mathbf{y}_{K}], \label{eq:Y=y1y2...yn}
\end{align}
where $f_{t}(\cdot; \theta_{t})$ denotes a pre-trained text feature extractor, such as the CLIP text encoder~\cite{clip}. The symbol $[\cdot]$ represents the concatenation operation, the matrix $\mathbf{Y} \in \mathbb{R}^{d \times K}$ is the collection of text features, with $d$ being the feature dimension. Note that text features have been normalized, \emph{i.e.}, $\forall i \in [1, K], \| \mathbf{y}_{i} \|=1$.

Second, we introduce two strategies of the CPG, namely, vanilla concept prototype and dual-level concept prototype. We will elucidate our approaches successively.

\paragraph{Vanilla Concept Prototype.}
We constructed a simplex ETF based on text features to maximize the separability between each pair of generated concept prototypes. Specifically, we perform a QR decomposition on $\mathbf{Y}$ to obtain the orthogonal basis of text features by:
\begin{equation}
  \mathbf{Y} = \mathbf{Q}\mathbf{R},
  \label{eq:y=qr}
\end{equation}
where $\mathbf{Q}=[\mathbf{q}_{1},\mathbf{q}_{2},\cdots,\mathbf{q}_{K}] \in \mathbb{R}^{d \times K}$ is an orthogonal matrix, $\mathbf{q}_{i}^{T}\mathbf{q}_{j} = 1$, when $i=j$, and 0 otherwise. $\mathbf{R} \in \mathbb{R}^{K \times K}$ is an upper triangular matrix. Then we can compute the vanilla concept prototypes following \cref{eq:ETF} by:
\begin{equation}
  \mathbf{E}=
    \sqrt{\frac{K}{K-1}}
    \mathbf{Q} 
    (\mathbf{I}_{K}-
    \frac{1}{K}
    \mathbf{1}_{K}\mathbf{1}_{K}^{T}),
  \label{eq:our_ETF}
\end{equation}
where $\mathbf{I}_{K}$ is a $K \times K$ identity matrix, and $\mathbf{1}_{K}$ represents a $K$-dimensional all-one vector. Besides, $\mathbf{E}=[\mathbf{e}_{1},\mathbf{e}_{2},\cdots,\mathbf{e}_{K}] \in \mathbb{R}^{d \times K}$ encompasses the concept prototypes for all classes, where $\mathbf{e}_{i} \in \mathbb{R}^{d}$ denotes the prototype for the $i$-th class. The similarity between pairwise prototypes can be expressed as:
\begin{equation}
  \mathbf{e}_{i}^{T} \mathbf{e}_{j} =
    \frac{K}{K-1}
    \mathbf{q}_{i}^{T} \mathbf{q}_{j}
    -\frac{1}{K-1}, 
  \forall i,j \in [1, K], 
  \label{eq:proto}\\
\end{equation}

\paragraph{Dual-level Concept Prototype.}
The construction of vanilla concept prototypes treats each class equally, without considering the similarity between classes. To enhance the distinction between similar categories, a direct idea is to perform more fine-grained differentiation for similar classes. Specifically, we set the similar classes as a group and construct dual-level concept prototypes, comprising coarse-grained prototypes for all groups and fine-grained prototypes for all classes within each group. We commence by computing the similarity matrix $\mathbf{S}$ between all classes as $\mathbf{S}=\mathbf{Y}^{T}\mathbf{Y}$. We use the matrix $\mathbf{S}$ to obtain the adjacency matrix $\mathbf{A}$ with the hyperparameter $p$ by:
\begin{equation}
  \mathbf{A}_{ij} = \{ \mathbf{S}_{ij} > p \}, \forall i,j \in [1,K], \label{eq:A=[S>p]}
\end{equation}
where $\{ \cdot \}$ denotes an Iverson bracket, a mathematical notation used to represent a logical value based on a condition. It equals 1 if the condition is true and 0 if false.

Subsequently, we construct a connectivity graph $\mathcal{G}=\{\mathcal{V},\mathcal{E}\}$, representing the relationships between all classes. $\mathcal{V}$ denotes the set of nodes (classes), and $|\mathcal{V}|=K$. $\mathcal{E}$ denotes the set of edges (whether the two classes are similar), and $|\mathcal{E}|=\sum_{i=1}^{K} \sum_{j=1}^{K}\mathbf{A}_{ij}$. To group similar nodes based on the adjacency matrix $\mathbf{A}$, we conduct a connectivity analysis on the graph $\mathcal{G}$ and get another graph $\mathcal{G}'$. Connectivity analysis refers to the process where if a path exists between two classes, they should be placed in the same group. The details can be found in the appendix. Please refer to the illustration in \cref{fig:framework}~(b) for a simple example.

Based on the above algorithms, we can group the text features $\mathbf{Y}=[\mathbf{Y}_{1},\mathbf{Y}_{2},\cdots,\mathbf{Y}_{N_{g}}]$, where $N_{g}$ is the number of groups. $\mathbf{Y}_{i} \in \mathbb{R}^{|g_{k}| \times d}$ represents the text features for the $i$-th group, and $|g_{k}|$ is the number of concepts in the $k$-th group. Then we compute the average text feature for each group by:
\begin{equation}
  \mathbf{\bar{Y}} = [
    \frac{1}{|g_{1}|}\sum_{i=1}^{|g_{1}|} \mathbf{Y}_{1,i}, 
    \frac{1}{|g_{2}|}\sum_{i=1}^{|g_{2}|} \mathbf{Y}_{2,i}, 
    \cdots,
    \frac{1}{|g_{N_{g}}|}\sum_{i=1}^{|g_{N_{g}}|} \mathbf{Y}_{N_{g},i}
  ].
  \label{eq:y_bar}
\end{equation}
Based on \cref{eq:y=qr,eq:our_ETF,eq:proto,eq:y_bar}, we can calculate the coarse-grained concept prototypes (CCP) through $\mathbf{\bar{Y}}$, denoted as:
\begin{equation}
  \mathbf{E}_{C} = [
    \mathbf{e}_{C,1}, 
    \mathbf{e}_{C,2}, 
    \cdots, 
    \mathbf{e}_{C,N_{g}}
  ] \in \mathbb{R}^{d \times N_{g}}.
  \label{eq:cp_c}
\end{equation}
Additionally, we can calculate the fine-grained concept prototypes (FCP) using $\mathbf{Y}_{i}$, represented as:
\begin{equation}
  \mathbf{E}_{F_{k}} = [
    \mathbf{e}_{F_{k},1}, 
    \mathbf{e}_{F_{k},2}, 
    \cdots, 
    \mathbf{e}_{F_{k},|g_{k}|}
  ] \in \mathbb{R}^{d \times |g_{k}|}.
  \label{eq:cp_f}
\end{equation}
Now we construct the dual-level concept prototypes for all classes. In other words, our CPG generates a coarse-grained prototype $\mathbf{e}_{C}$ and a fine-grained prototype $\mathbf{e}_{F}$ for each class.

\subsection{Coarse-to-Fine Calibrator}
The coarse-to-fine calibrator (C2F) is proposed to align the image features with the corresponding prototypes. Given an input image $x$ and its class name $y$, we extract features of the image $x$ as $\mathbf{x}=f_{i}(x;\theta_{i}) \in \mathbb{R}^{d}$, where $f_{i}(\cdot;\theta_{i})$ is a pre-trained model, such as ViT~\cite{vit} or CLIP image encoder~\cite{clip}. Assuming $y$ is the $j$-th class from the $i$-th group, its concept prototypes are referred to as $\mathbf{e}_{C,i}$ and $\mathbf{e}_{F_{i},j}$.
The C2F module consists of the coarse-grained layer $g_{C}(\cdot;\varphi_{C})$ and the fine-grained layers $g_{F_{k}}(\cdot;\varphi_{F_{k}})$. The coarse-grained feature $\mathbf{x}_{C}$ and fine-grained features $\mathbf{x}_{F}$ can be computed by:
\begin{equation}
\begin{aligned}
  \mathbf{x}_{C} &= g_{C}(\mathbf{x};\varphi_{C}) \in \mathbb{R}^{d}, \\
  \mathbf{x}_{F} &= g_{F_{i}}([\mathbf{x}, \mathbf{x}_{C}];\varphi_{F}) \in \mathbb{R}^{d}.
\end{aligned}
\label{eq:c2f}
\end{equation}

\paragraph{Dual Dot-Regression Loss.}
To train the C2F module, we propose the Dual Dot-Regression (DDR) loss function, denoted as $\mathcal{L}$. We train the parameter sets $\varphi_{C}$ and $\varphi_{F}$ by minimize the DDR loss with a hyperparameter $\alpha$:
\begin{equation}
\begin{aligned}
  \min_{\varphi_{C}, \varphi_{F}}&
  \mathcal{L} (\mathbf{x}_{C}, \mathbf{x}_{F}, \mathbf{e}_{C,i}, \mathbf{e}_{F_{i},j}) = \\
  &\alpha (\mathbf{x}_{C}^{T} \mathbf{e}_{C,i} - 1)^{2} + 
  (1-\alpha) (\mathbf{x}_{F}^{T} \mathbf{e}_{F_{i},j} - 1)^{2},
\end{aligned}
\label{eq:ddrloss}
\end{equation}

\subsection{Theoretical Analysis}
\paragraph{Theorem 1.} \emph{The angle between any pair of CCPs or FCPs is larger than or equal to the angle between any pair of vanilla concept prototypes}:
\begin{equation}
  \left.
  \begin{array}{r}
    \langle \mathbf{e}_{C,m},\mathbf{e}_{C,n} \rangle \geq \langle \mathbf{e}_{i},\mathbf{e}_{j} \rangle,\\
    \langle \mathbf{e}_{F_{k},m},\mathbf{e}_{F_{k},n} \rangle \geq \langle \mathbf{e}_{i},\mathbf{e}_{j} \rangle, \forall k,
  \end{array}
  \right\}\forall i \neq j, m \neq n.
\end{equation}
This theorem demonstrates that our proposed dual-level concept prototype has larger inter-class angles, indicating better classification capability than the vanilla concept prototype. The proof of the theorem is provided in the appendix.

%%%%%%%%%%%%%%%%%%%%%%%%%%%%%%%

\section{Experiments}
\subsection{Experimental Settings}
\paragraph{Datasets.} We conducted experiments on three multi-domain datasets, include DomainNet~\cite{peng2019moment}, CDDB~\cite{li2023continual}, and CORe50~\cite{lomonaco2017core50}.
\textbf{DomainNet} emerges as a large-scale dataset for DIL and domain adaptation, whose images are sourced from six domains marked by prominent inter-domain variations, with each domain including 345 categories. The training set of DomainNet consists of 409,832 images, while the test set comprises 176,743 images.
\textbf{CORe50} is an object recognition dataset involving 11 distinct domains (50 classes per domain), with 8 domains for training and 3 for testing.
\textbf{CDDB} is specifically crafted for deepfake detection, encompassing 12 distinct deepfake methodologies and 3 different evaluation scenarios. We opt for the most challenging HARD track as suggested by S-Prompts~\cite{wang2022s}.

\paragraph{Evaluation Metrics.}
There are three commonly used evaluation metrics for DIL: (1) the average accuracy ($A_{T}$) at the end of training on all $T$ domains; (2) the forgetting degree ($F_{T}$) following~\cite{li2023continual}, and the formulas for calculating $A_{T}$ and $F_{T}$ are detailed in the appendix; (3) the ``Buffer'' represents additional data stored by the model for incremental learning. This data may include images from old domains used for rehearsal.

\begin{table}[t]
  \small
  \centering
  \setlength{\tabcolsep}{2pt}
  \begin{tabular}{lccc}
    \toprule
    Method & Buffer$(\downarrow)$ & $A_{T}(\uparrow)$ & $F_{T}(\uparrow)$ \\
    \midrule
    DyTox~\cite{douillard2022dytox} & \multirow{3}{*}{50/class} & 62.94 & - \\
    DARE~\citep{jeeveswaran2024gradual} & & 32.32* & -22.98 \\
    DARE++~\cite{jeeveswaran2024gradual} & & 40.51* & - \\
    \midrule
    EWC~\cite{kirkpatrick2017overcoming} & \multirow{13}{*}{0/class}  & 47.62 & - \\
    LwF~\cite{li2017learning} &  & 49.19 & -5.01 \\
    SimCLR~\cite{chen2020simple} &  & 44.20 & - \\
    BYOL~\cite{grill2020bootstrap} &  & 49.70 & - \\
    Barlow Twins~\cite{zbontar2021barlow} &  &  48.90 & - \\
    SupCon~\cite{khosla2020supervised} &  & 50.90 & - \\
    L2P~\cite{wang2022learning} &  & 40.15$\dag$ & -2.25 \\
    DualPrompt~\cite{wang2022dualprompt} &  & 43.79$\dag$ & \underline{-2.03} \\
    S-iP~\cite{wang2022s} &  & 50.62$\dag$ & -2.85\\
    CODA-P~\cite{smith2023coda} &  & 47.42$\dag$ & -3.46 \\
    C-Prompt~\cite{liu2024compositional} & & \underline{58.68}$\dag$ & - \\
    \rowcolor{gray!20}
    DualCP (ours) & & \textbf{60.13}$\dag$ & \textbf{-1.96} \\
    \bottomrule
  \end{tabular}
  \caption{\textbf{Experimental results on the DomainNet dataset.} 
  $\dag$ denotes that the method is based on the pre-trained ViT-B/16 model.
  * denotes that DARE is based on ResNet-18. The best result within rehearsal-free methods is indicated by \textbf{bold}, and the second is marked by \underline{underline}.}
  \label{tab:domainnet}
\end{table}

\begin{table}[t]
  \setlength{\tabcolsep}{3pt}
  \small
  \centering
  \begin{tabular}{lccc}
    \toprule
    Method & Buffer$(\downarrow)$ & $A_{T}(\uparrow)$ & $F_{T}(\uparrow)$\\
    \midrule
    LRCIL~\cite{pellegrini2020latent} & \multirow{3}{*}{100/class}  & 76.39 & -4.39 \\
    iCaRL~\cite{marra2019incremental} &  & 79.76 & -8.73 \\
    LUCIR~\cite{hou2019learning} &  & 82.53 & -5.34 \\
    \midrule
    LRCIL~\cite{pellegrini2020latent} & \multirow{4}{*}{50/class}  & 74.01 & -8.62 \\
    iCaRL~\cite{marra2019incremental} &  & 73.98 & -14.50 \\
    LUCIR~\cite{hou2019learning} &  & 80.77 & -7.85 \\
    DyTox~\cite{douillard2022dytox} &  & 86.21 & -1.55 \\
    \midrule
    EWC~\cite{kirkpatrick2017overcoming} & \multirow{8}{*}{0/class} &  50.59 & -42.62 \\
    LwF~\cite{li2017learning} & &  60.94 & -13.53 \\
    DyTox~\cite{douillard2022dytox} & &  51.27 & -45.85 \\
    L2P~\cite{wang2022learning} & &  61.28$\dag$ & -9.23 \\
    DualPrompt~\cite{wang2022dualprompt} & & 64.80$\dag$ & -8.74 \\
    S-iP~\cite{wang2022s} & & \underline{74.51}$\dag$ & \underline{-1.30} \\
    CODA-P~\cite{smith2023coda} & & 70.54$\dag$ & -5.53 \\
    \rowcolor{gray!20}
    DualCP (ours) &  & \textbf{82.16}$\dag$ & \textbf{-0.73} \\
    \bottomrule
  \end{tabular}
  \caption{\textbf{Experimental results on the Hard track of CDDB.} $\dag$ denotes that the method is based on the pre-trained ViT-B/16 model. The best result within rehearsal-free methods is indicated by \textbf{bold}, and the second is marked by \underline{underline}.}
  \label{tab:cddb}
\end{table}

\begin{table}[t]
  \small
  \centering
  \setlength{\tabcolsep}{3pt}
  \begin{tabular}{lccc}
    \toprule
    Method & Buffer$(\downarrow)$ & $A_{T}(\uparrow)$ \\
    \midrule
    ER~\cite{chaudhry2019tiny} & \multirow{7}{*}{50/class} &  80.10 \\
    GDumb~\cite{prabhu2020gdumb} & & 74.92 \\
    BiC~\cite{wu2019large} & & 79.28 \\
    DER++~\cite{buzzega2020dark} & & 79.70 \\
    Co$^{2}$L~\cite{cha2021co2l} & & 79.75 \\
    DyTox~\cite{douillard2022dytox} & & 79.21 \\
    L2P~\cite{wang2022learning} & & 81.07 \\
    \midrule
    EWC~\cite{kirkpatrick2017overcoming} & \multirow{9}{*}{0/class} & 74.82 \\
    LwF~\cite{li2017learning} & & 75.45 \\
    L2P~\cite{wang2022learning} & & 78.33$\dag$ \\
    DualPrompt~\cite{wang2022dualprompt} &  & 80.25$\dag$ \\
    S-iP~\cite{wang2022s} & & 83.13$\dag$ \\
    CODA-P~\cite{smith2023coda} & & \underline{85.68}$\dag$ \\
    C-Prompt~\cite{liu2024compositional} & & 85.31$\dag$ \\
    \rowcolor{gray!20}
    DualCP (ours) & & \textbf{88.10}$\dag$ \\
    \bottomrule
  \end{tabular}
  \caption{\textbf{Experimental results on the CORe50 dataset.} Note that $F_{T}$ is not applicable to CORe50 because the training and test domains do not overlap. $\dag$ denotes that the method is based on the pre-trained ViT-B/16 model. The best result within rehearsal-free methods is indicated by \textbf{bold}, and the second is marked by \underline{underline}.}
  \label{tab:core50}
\end{table}

\paragraph{Implementation Details.}
We employ an SGD optimizer with an initial learning rate of 0.1 and a cosine decay schedule.
Additionally, we apply the weight decay of $2e^{-4}$ for regularization to mitigate overfitting.
The training consists of 20 epochs on all datasets except DomainNet, which extends to 30 epochs.
The mini-batch size is set to 128.
The hyperparameters $p$ and $\alpha$ are set to 0.85 and 0.5, respectively. The ablations of $p$ and $\alpha$ are provided in the appendix.

\subsection{Main Results}

\paragraph{Baselines.}
The current DIL methods can be broadly categorized into rehearsal-based and rehearsal-free methods.
\textbf{Rehearsal-based} methods select and retrain a subset of images as exemplars of the domain when training. 
Representative methods include ER, LRCIL, iCaRL, and LUCIR.
\textbf{Rehearsal-free} methods do not require saving images from the old domain. Representative methods include EWC, LwF, DyTox, L2P, S-Prompts, \emph{etc}.
Note that rehearsal-based methods often require storing thousands of images, ranging in size \emph{from 100MB to 3GB}. In contrast, rehearsal-free methods may require only a small amount of learnable parameters, occupying \emph{1-50MB} of space. Therefore, rehearsal-free methods significantly outperform rehearsal-based methods in terms of storage space requirements.
The proposed DualCP belongs to the rehearsal-free setting, so we prioritize comparing it with similar methods. Additionally, we also list the state-of-the-art rehearsal-based methods for reference.

\paragraph{Comparison with State-of-the-arts.}

We compare our approach with other state-of-the-art (SOTA) methods on three DIL benchmark datasets. The methods are grouped based on the number of images per class to be retained, with ``0/class'' indicating a rehearsal-free approach.

\cref{tab:domainnet} illustrates the comparison results on the DomainNet dataset. Our DualCP surpasses the SOTA method C-Prompt (60.13\% \emph{vs.} 58.68\%), even coming close to the performance of the DyTox method that utilizes rehearsal. Besides, our method achieved the best performance in preventing forgetting among all methods (-1.96\%).
\cref{tab:cddb} showcases results on CDDB, where our DualCP surpasses the best rehearsal-free method by a large margin (82.46\% \emph{vs.} 74.51\%), and is comparable to LUCIR, which utilizes a substantial buffer (``100/class''). It falls just behind the rehearsal-based DyTox method. Additionally, our method outperforms the SOTA method in mitigating forgetting (-0.73\% \emph{vs.} -1.30\%).
\cref{tab:core50} presents the comparisons on CORe50, revealing that our method achieves the best results in both rehearsal-based (88.10\% \emph{vs.} 81.07\%) and rehearsal-free (88.10\% \emph{vs.} 85.68\%) tracks.

\subsection{Ablation Study}

\begin{table}[t]
  \small
  \centering
  \setlength{\tabcolsep}{2.0pt}
  \begin{tabular}{cccccc}
    \toprule
    \multirow{3}{*}{\diagbox[width=2.3cm]{Methods}{Datasets}} & \multicolumn{2}{c}{DomainNet} & \multicolumn{2}{c}{CDDB} & CORe50 \\
         & \multicolumn{2}{c}{(345 classes)} & \multicolumn{2}{c}{(2 classes)} & (50 classes) \\
         & $A_{T}(\uparrow)$ & $F_{T}(\uparrow)$ & $A_{T}(\uparrow)$ & $F_{T}(\uparrow)$ & $A_{T}(\uparrow)$ \\
    \midrule
    VanillaCP & 56.22 & -2.80 & \textbf{82.16} & \textbf{-0.73} & 86.27 \\
    DualCP (proposed) & \textbf{60.13} & \textbf{-1.96} & \textbf{82.16}* & \textbf{-0.73}* & \textbf{88.10} \\
    \bottomrule
  \end{tabular}
  \caption{\textbf{Comparison of different concept prototype designs in our framework.} * represents that CDDB only contains two classes: real and fake, thus it cannot be divided into more groups and cannot apply to the DualCP.}
  \label{tab:dualcp}
\end{table}

\paragraph{Effectiveness of Dual Concept Prototype.}
We introduce two solutions to generate the concept prototypes: a single-level concept prototype (VanillaCP) and a dual-level concept prototype (DualCP). \cref{tab:dualcp} demonstrates a consistent improvement of DualCP over VanillaCP, especially on datasets containing a large number of classes. This indicates that our DualCP contributes to distinguishing similar classes.

\begin{table*}[t]
  \small
  \centering
\begin{tabular}{l|cc|cc|ccccc}
\hline
  \multirow{2}{*}{Methods} & \multicolumn{2}{c|}{Image Feature Extractor} & \multicolumn{2}{c|}{CPG Guidance} 
           & \multicolumn{2}{c}{DomainNet} & \multicolumn{2}{c}{CDDB} & CORe50 \\
           & ViT-B/16 & CLIP-Image & ViT-B/16-BD & CLIP-Text & $A_{T}(\uparrow)$ & $F_{T}(\uparrow)$ & $A_{T}(\uparrow)$ & $F_{T}(\uparrow)$ & $A_{T}(\uparrow)$ \\
  \hline
  \multirow{2}{*}{S-Prompts}
           & \ding{51} &  &  &          & 50.62 & -2.85 & 74.51 & -1.30 & 83.13 \\
           & & \ding{51} & & \ding{51}* & 67.78 & -1.64 & 88.65 & -0.69 & 89.06 \\
  MoP-CLIP
           & & \ding{51} & & \ding{51}* & 69.70 & -       & 88.54 & -0.79 & 92.29 \\
  \hdashline
  \multirow{4}{*}{DualCP (ours)} 
         & \ding{51} & & \ding{51} & & 60.13 & -1.96 & 82.16 & -0.73 & 88.10 \\
         & \ding{51} & & & \ding{51} & 62.73 & -1.81 & 83.05 & -0.76 & 88.55 \\
         & & \ding{51} & \ding{51} & & 69.31 & -1.49 & 91.86 & -0.35 & 89.98 \\
         & & \ding{51} & & \ding{51} & \textbf{72.46} & \textbf{-1.26} & \textbf{92.34} & \textbf{-0.32} & \textbf{90.59} \\
  \hline
  \end{tabular}
  \caption{\textbf{Comparison on different backbones.} ViT-B/16-BD means that image features extracted from the base domain by ViT-B/16 are used to guide the CPG module. * indicates that S-Prompts and MoP-CLIP utilize CLIP text encoder but are unrelated to the CPG module. The proposed CPG module is only used for our DualCP.}
  \label{tab:backbone}
\end{table*}

\paragraph{Comparsion on Different Backbones.}
To validate the generality of our approach, we conducted experiments on both ViT and CLIP, a multimodal model based on contrastive learning with an image encoder and a text encoder. \cref{tab:backbone} presents the accuracy of our DualCP under different settings, as well as that of S-Prompts and MoP-CLIP.

\textbf{Image Feature Extractor.} We extract image features using two optional setting, \emph{i.e.}, ViT-B/16 or CLIP image encoder, as the feature extractor ($f_{i}(\cdot;\theta_{i})$).

\textbf{CPG Guidance.} We proposed two methods to extract the concept prototype in the CPG module. One is based on the CLIP text encoder, as shown in \cref{eq:text_encoder}. When CLIP is unavailable, we introduce an alternative approach using ViT. We use ViT-B/16 to extract image features from the base domain, and then calculate the mean features $\mathbf{y}'_{i}$ for each class. We use $\mathbf{Y}'=\{\mathbf{y}'_{1}, \mathbf{y}'_{2}, \cdots, \mathbf{y}'_{k}\}$ to guide the construction of the concept prototype, as shown in \cref{eq:y=qr}.

\begin{figure}[t]
  \centering
  \includegraphics[width=0.48\textwidth]{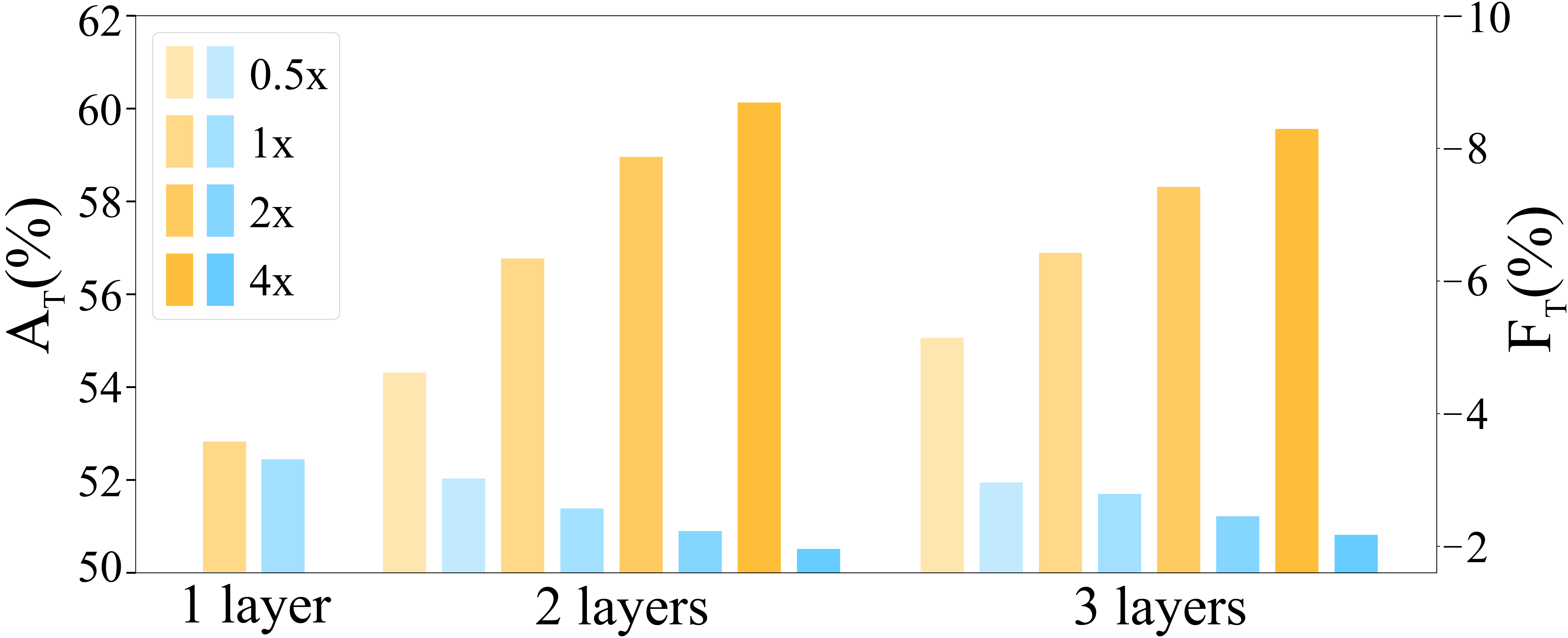}
  \caption{\textbf{Ablation study of the C2F module on DomainNet dataset.} $A_{T}$, $F_{T}$ denotes the average accuracy and the forgetting degree, respectively. The hidden dimensions are set as multiples of the image feature dimensions of 768 in the ViT-B/16 backbone. ``0.5x, 1x, 2x, 4x'' correspond to hidden dimensions of 384, 768, 1536, and 3072, respectively.}
  \label{fig:ab_c2f}
\end{figure}

\paragraph{Ablations of C2F Design.}
Our C2F comprises a coarse-grained layer $g_{C}$ and multiple fine-grained layers $g_{F_{i}}$, which are implemented as a multi-layer perceptron (MLP). As depicted in \cref{fig:ab_c2f}, we conduct ablation experiments on the number of layers and the hidden dimension of our C2F to assess the influence of MLP on model performance.

\begin{figure}[t]
  \centering
  \includegraphics[width=0.39\textwidth]{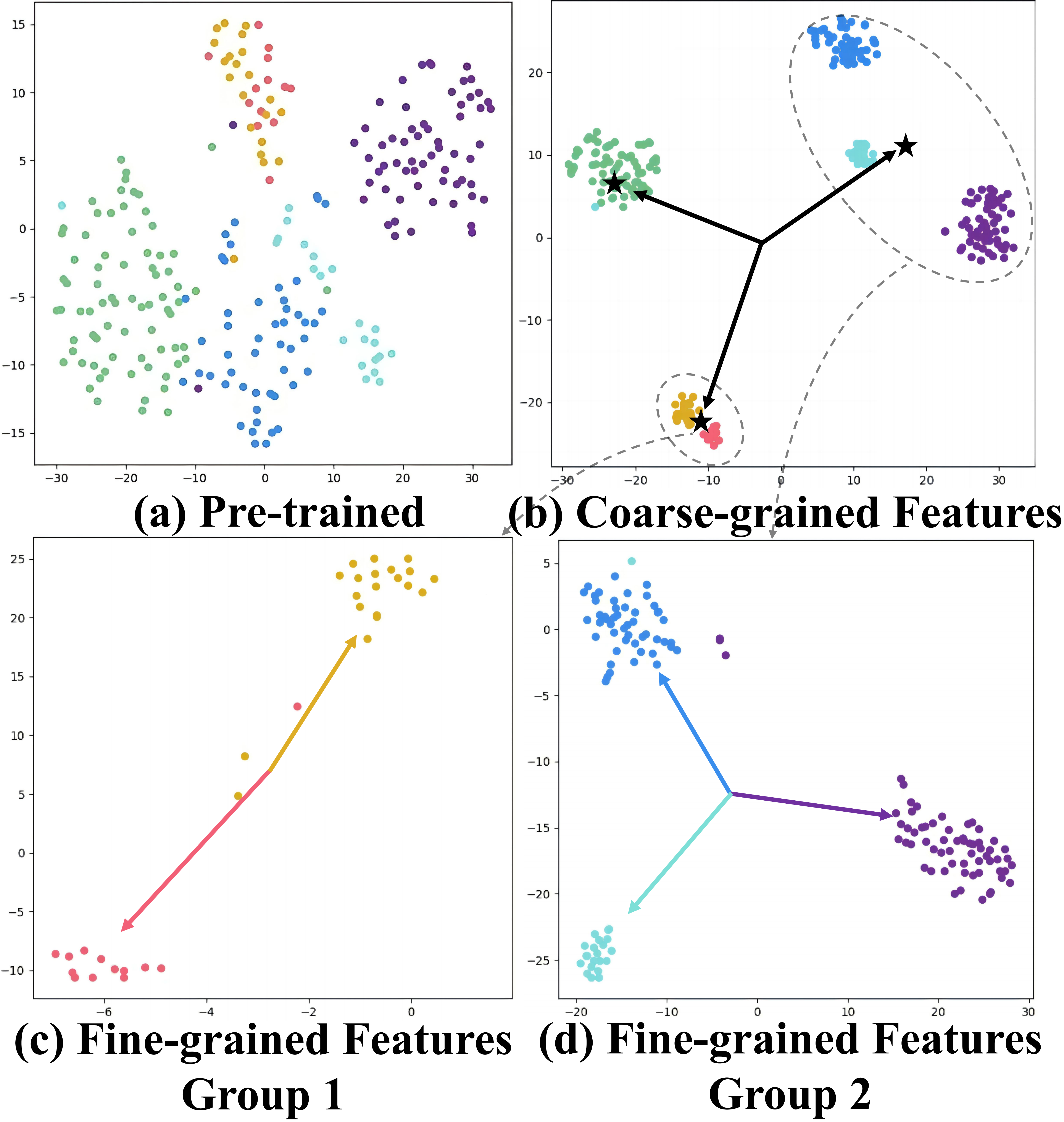}
  \caption{\textbf{t-SNE visualization} of feature space for common classes in DomainNet, with flower represented in green, cat and dog in (c), and boat, bicycle, and bus in (d). The pentagrams represent the average image features of a group.}
  \label{fig:tsne}
\end{figure}

\subsection{Visualization}
We utilize t-SNE visualization to demonstrate the effectiveness of our DualCP. We choose six common classes (cat, flower, dog, boat, bicycle, and bus) from DomainNet for presentation. As shown in \cref{fig:tsne}, (a) represents the image features extracted by the pre-trained ViT model. (b-d) depict the image features extracted by our DualCP. Our model categorizes similar classes into the same group based on semantics, such as cat and dog. (b) represents the coarse-grained features extracted by DualCP. (c) and (d) represent fine-grained features of different groups. As illustrated in \cref{fig:tsne} (c)(d), our dual-level concept prototypes effectively distinguish similar classes.

\section{Conclusion and Future Works}

This paper introduces a novel approach to RFDIL inspired by humans' incremental cognitive processes. We proposed constructing dual-level concept prototypes (DualCP) for each class across domains to address the zero-sum game between learning new domains and preserving old ones. By aligning features from different domains to the same feature space, we avoid compromising the feature space of old domains while accommodating new domain features. Extensive experiments on three datasets with different backbones consistently show that our DualCP outperforms existing SOTA methods.
Furthermore, our method is expected to be extended to other applications, such as domain-incremental object detection. It may also provide a reference for the development of generalized neural collapse.

\section{Acknowledgments}
This work was funded by the National Natural Science Foundation of China under Grant No.U21B2048 and No.62302382, Shenzhen Key Technical Projects under Grant CJGJZD2022051714160501, China Postdoctoral Science Foundation No.2024M752584, and Natural Science Foundation of Shaanxi Province No.2024JC-YBQN-0637.

\bibliography{aaai25}

\clearpage
\appendix
\section{Appendix}

\subsection{Further Details of the Training and Inference}
\cref{alg:framework} introduces how to train and test our Dual-level Concept Prototype (DualCP) framework. The symbols correspond to those used in the Problem Formulation subsection and the Method section.

\begin{algorithm}[ht]
    \caption{Dual-level Concept Prototype Algorithm}
    \label{alg:framework}
    \textbf{Input}: Training set $\mathcal{X}_{1},\cdots,\mathcal{X}_{T}$, test set $\mathcal{Z}_{1},\cdots,\mathcal{Z}_{T}$, and the set of labels $\mathcal{C}$. \\
    \textbf{Output}: The predicted labels of test images.
    
    \begin{algorithmic}[1] %[1] enables line numbers
    \STATE Compute the text features $\mathbf{Y}$ of $\mathcal{C}$ using \cref{eq:text_encoder,eq:Y=y1y2...yn}.
    \STATE Compute the coarse-grained concept prototype $\mathbf{E}_{C}=[\mathbf{e}_{C,1},\cdots,\mathbf{e}_{C,N_{g}}]$ and the fine-grained concept prototype $\mathbf{E}_{F_{k}}=[\mathbf{e}_{F_{k},1},\cdots,\mathbf{e}_{F_{k},|g_{k}|}]$ using \cref{eq:y=qr,eq:proto,eq:cp_c,eq:cp_f}.
    \FOR {$t$ in $[1,T]$}
        \FOR {i in $[1,N_{t}]$}
            \STATE Compute the image features $\mathbf{x}_{t,i}=f_{i}(x_{t,i};\theta_{i})$.
            \STATE Compute $\mathbf{x}_{C}^{(t)}$ and $\mathbf{x}_{F}^{(t)}$ by the C2F using \cref{eq:c2f}.
            \STATE Optimize $\varphi_{C}^{(t)}$ and $\varphi_{F}^{(t)}$ by minimizing $\mathcal{L}$ using \cref{eq:ddrloss}.
        \ENDFOR
        \STATE Compute the average features $\mathbf{x}_{t}=\frac{1}{N_{t}} \sum_{i=1}^{N_{t}} \mathbf{x}_{t,i}$.
    \ENDFOR

    \STATE Get the test set $\mathcal{Z}_{1 \sim T} = \bigcup_{\tau = 1}^{T} \mathcal{Z}_{\tau}$
    \FOR {$z$ in $\mathcal{Z}_{1 \sim T}$}
        \STATE Compute the image features $\mathbf{z}=f_{i}(z;\theta_{i})$.
        \STATE Compute the domain label $p = \underset{1 \le t \le T}{\arg\max} (\frac{\mathbf{z} \cdot \mathbf{x}_{t}}{\|\mathbf{z}\| \|\mathbf{x}_{t}\|})$.
        \STATE Compute $\mathbf{z}_{C}=g_{C}(\mathbf{z};\varphi_{C}^{(p)})$ and $\mathbf{z}_{F}=g_{F}(\mathbf{z};\varphi_{F}^{(p)})$.
        \STATE The coarse-grained label $c' = \underset{1 \le i \le N_{g}}{\arg\max} \frac{\mathbf{z}_{C} \cdot \mathbf{e}_{C,i}}{\|\mathbf{z}_{C}\| \|\mathbf{e}_{C,i}\|}$,
        \STATE The fine-grained label $f' = \underset{1 \le i \le |g_{c'}|}{\arg\max} \frac{\mathbf{z}_{F} \cdot \mathbf{e}_{F_{C'},i}}{\|\mathbf{z}_{F}\|\|\mathbf{e}_{F_{C'},i}\|}$.
        \STATE Output the predicted label $c_{f}=f'+\sum_{i=1}^{c'-1}|g_{C'}|$.
    \ENDFOR
    \end{algorithmic}
\end{algorithm}

\subsection{Details of Connectivity Analysis Algorithms}
\cref{alg:dfs} describes the depth-first search algorithm, which explores connected nodes for a given node in a graph. \cref{alg:connect} describes the connectivity analysis algorithm, which traverses all nodes in the given graph and ensures that there is an edge between every pair of nodes that have a connected path.

\begin{algorithm}[ht]
    \caption{Depth-First Search (DFS) to get the nodes of the same group.}
    \label{alg:dfs}
    \textbf{Global variable}: list $g$ to store the nodes of the same group. \\
    \textbf{Input}: adjacency matrix $\mathbf{A}$, array $v$ for storing node visitation status, node $n$ that need to be grouped. \\
    \textbf{Output}: list $g$.
    
    \begin{algorithmic}[1] %[1] enables line numbers
        \IF {$g$ is None}
            \STATE Initialize an empty list $g$.
        \ENDIF
        \STATE Add node $n$ to list $g$.
        \FOR {$n',c$ in enumerate($\mathbf{A}[n])$}
            \STATE \# Variable $c$ means the connection status between $n$ and $n'$, \emph{i.e.}, $c=1$ indicates a connection, while $c=0$ indicates no connection.
            \STATE \# Node $n'$ is the neighbor of the node $n$ if $c=1$.
            \IF {$c = 1$ \AND $v[n'] = 0$}
                \STATE $g\leftarrow$ DFS($\mathbf{A},v,n',g$)
            \ENDIF
        \ENDFOR
        \RETURN $g$
    \end{algorithmic}
\end{algorithm}

\begin{algorithm}[ht]
    \caption{Connectivity Analysis via DFS}
    \label{alg:connect}
    \textbf{Input}: adjacency matrix $\mathbf{A}$, node-set $\mathcal{V}$, number of nodes $K$, \cref{alg:dfs}, \emph{i.e.}, DFS($\cdot,\cdot,\cdot,\cdot$).\\
    \textbf{Output}: list $G'$ to store the lists of groups.
    
    \begin{algorithmic}[1] %[1] enables line numbers
        \STATE Initialize an array $v$ of length $K$ for storing node visitation status.
        \STATE Initialize an empty list $G'$.
        \FOR {$i=0$ to $K-1$}
            \STATE $v[i] = 0$.
        \ENDFOR
        
        \FOR {$n$ \textbf{in} $ \mathcal{V} $}
            \IF {$v[n] = 0$}
            \STATE Initialize an empty list $g$ to store the nodes in the same group.
            \STATE $g\leftarrow$ DFS($\mathbf{A},v,n,g$)
            \STATE Add list $g$ to list $G'$.
            \STATE $v[n] = 1$.
            \ENDIF
        \ENDFOR
        \RETURN $G'$
    \end{algorithmic}
\end{algorithm}

\subsection{The Proof of Theorem 1}
\paragraph{Theorem 1.} \emph{The angle between any pair of CCPs or FCPs is larger than or equal to the angle between any pair of vanilla concept prototypes}:
\begin{equation}
  \left.
  \begin{array}{r}
    \langle \mathbf{e}_{C,m},\mathbf{e}_{C,n} \rangle \geq \langle \mathbf{e}_{i},\mathbf{e}_{j} \rangle,\\
    \langle \mathbf{e}_{F_{k},m},\mathbf{e}_{F_{k},n} \rangle \geq \langle \mathbf{e}_{i},\mathbf{e}_{j} \rangle, \forall k,
  \end{array}
  \right\}\forall i \neq j, m \neq n.
\end{equation}

\paragraph{Proof.} For a classification task with $K$ classes, we divide them into $N_{g}$ groups, where the $k$-th group contains $|g_{k}|$ classes, \emph{i.e.},
\begin{equation}
    K=\sum_{k=1}^{N_{g}} |g_{k}|.
    \label{eq:numgroup}
\end{equation}
Following the vanilla concept prototypes algorithm, we construct $\mathbf{E}=[\mathbf{e}_{1},\mathbf{e}_{2},\cdots,\mathbf{e}_{K}]$. Following the proposed dual-level concept prototype algorithm, we construct the coarse-grained concept prototypes (CCPs) as $\mathbf{E}_{C}=[\mathbf{e}_{C,1},\mathbf{e}_{C,2},\cdots,\mathbf{e}_{C,N_{g}}]$ and the fine-grained concept prototypes (FCPs) as $\mathbf{E}_{F_{k}}=[\mathbf{e}_{F_{k},1},\mathbf{e}_{F_{k},2},\cdots,\mathbf{e}_{F_{k},|g_{k}|}]$. Following \cref{eq:proto}, we have:
\begin{equation}
  \left.
  \begin{array}{r}
    \mathbf{e}_{i}^{T}\mathbf{e}_{j}=-\frac{1}{K-1}, \\
    \mathbf{e}_{C,i}^{T}\mathbf{e}_{C,j}=-\frac{1}{N_{g}-1},\\
    \mathbf{e}_{F_{k},i}^{T}\mathbf{e}_{F_{k},j}=-\frac{1}{|g_{k}|-1},
  \end{array}
  \right\}\forall i \neq j.
\end{equation}
Then, we have:
\begin{equation}
  \begin{aligned}
    \mathbf{e}_{C,m}^{T}\mathbf{e}_{C,n}-\mathbf{e}_{i}^{T}\mathbf{e}_{j} & = -\frac{1}{N_{g}-1}-(-\frac{1}{K-1}) \\
    & = \frac{N_{g}-K}{(N_{g}-1)(K-1)}
  \end{aligned}
\end{equation}
Following \cref{eq:numgroup}, we have $1 \leq N_{g} \leq K$ and $1 \leq |g_{k}| \leq K, \forall k$. Then we have $\mathbf{e}_{C,m}^{T}\mathbf{e}_{C,n}-\mathbf{e}_{i}^{T}\mathbf{e}_{j} \leq 0$, \emph{i.e.}, $\mathbf{e}_{C,m}^{T}\mathbf{e}_{C,n} \leq \mathbf{e}_{i}^{T}\mathbf{e}_{j}$.
Given that $\cos\langle \mathbf{a}, \mathbf{b} \rangle = \mathbf{a}^{T}\mathbf{b}$ and $\cos(\cdot)$ is a decreasing function when $\langle \mathbf{a}, \mathbf{b} \rangle \in (0, \pi)$, therefore:
\begin{equation}
    \cos \langle \mathbf{e}_{C,m},\mathbf{e}_{C,n} \rangle \leq \cos \langle \mathbf{e}_{i},\mathbf{e}_{j} \rangle,
\end{equation}
\begin{equation}
    \langle \mathbf{e}_{C,m},\mathbf{e}_{C,n} \rangle \geq \langle \mathbf{e}_{i},\mathbf{e}_{j} \rangle.
\end{equation}
Similarly, it can be proven that $\langle \mathbf{e}_{F_{k},m},\mathbf{e}_{F_{k},n} \rangle \geq \langle \mathbf{e}_{i},\mathbf{e}_{j} \rangle, \forall k$.

\subsection{Details of Evaluation Metrics}
For an incremental setting with $T$ domains, assuming $B \in \mathbb{R}^{T \times T}$ is an upper triangular matrix where $B_{i,j}$ denotes the test accuracy of the $i$-th domain after training on the $j$-th domain ($j \ge i$). $A_{t}$ denotes the accuracy tested on $\mathcal{Z}_{1 \sim t}$ after training the model sequentially on $\{\mathcal{X}_{1}, \mathcal{X}_{2}, \cdots, \mathcal{X}_{t}\}$. $A_{T}$ refers to the model's average accuracy tested on $\mathcal{Z}_{1 \sim T}$ after completing training on all $T$ domains in the benchmark dataset. The average accuracy ($A_{T}$) can be calculated by:
\begin{equation}
  A_{T} = \frac{1}{T} \sum_{i=1}^{T} B_{i,T}.
\end{equation}
Besides, We compute the forgetting degree ($F_{T}$) following~\cite{li2023continual} as:
\begin{align}
    BWT_{i}&=\frac{1}{T-i} \sum_{j=i+1}^{T} (B_{i,j}-B_{i,i}), \\
    F_{T}&=\frac{1}{T-1} \sum_{i=1}^{T-1} BWT_{i},
\end{align}
where $BWT_{i}$ represents the mean of backward transfer degradation.

\begin{figure}[t]
  \centering
  \includegraphics[width=0.45\textwidth]{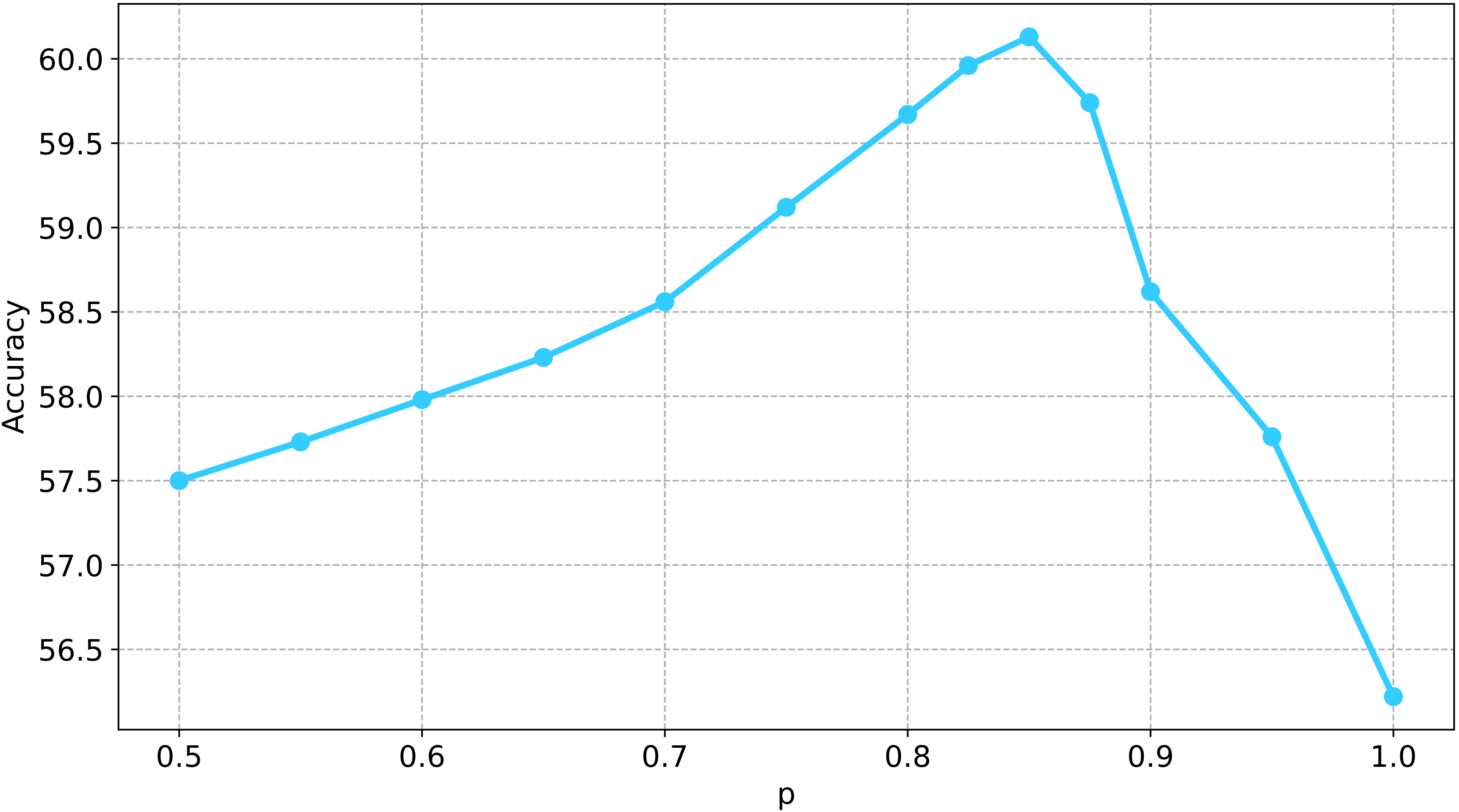}
  \caption{\textbf{Ablation study of the hyperparameter $p$ on the DomainNet dataset.} When the hyperparameter $p$ is set to 1, it implies that each group contains only one class. In this case, our DualCP method degenerates into VanillaCP.}
  \label{fig:ab_p}
\end{figure}

\begin{table}[t]
  \small
  \centering
  \begin{tabular}{lccccc}
    \toprule
    \multicolumn{1}{c}{$\alpha$} & 0.1   & 0.25  & 0.5   & 0.75  & 0.9   \\
    \midrule
    $A_{T}(\uparrow)$  & 58.92 & 59.75 & \textbf{60.13} & 59.32 & 58.06 \\
    $F_{T}(\uparrow)$  & -2.84 & -2.30  & \textbf{-1.96} & -2.38 & -3.02 \\
    \bottomrule
    \end{tabular}
  \caption{Ablation study of the hyperparameter $\alpha$ in our DDR loss function on the DomainNet dataset.}
  \label{tab:ab_loss}
\end{table}

\subsection{More Ablations of the Hyperparameters}
As shown in \cref{fig:ab_p}, we conducted an ablation study on the hyperparameter $p$. Our DualCP achieved the best results when $p$ was set to 0.85. Additionally, we performed an ablation study on the hyperparameter $\alpha$ to balance the contributions of different components in the DDR loss function. As illustrated in \cref{tab:ab_loss}. our method achieves the highest accuracy when $\alpha$ is set to 0.5, where coarse-grained and fine-grained features are optimized with equal weights.

\end{document}